\documentclass{article}

\usepackage[nonatbib,final]{neurips_2025_ml4ps}

\usepackage[
  backend=biber,
  style=numeric,
  maxnames=10,
  minnames=1,
  sorting=nyt,
  url=true,
  doi=true,
  eprint=false
]{biblatex}
\renewbibmacro*{url+urldate}{%
  \printfield{url}%
}

\usepackage[utf8]{inputenc} 
\usepackage[T1]{fontenc}    
\usepackage{hyperref}       
\usepackage{url}            
\usepackage{booktabs}       
\usepackage{amsfonts}       
\usepackage{nicefrac}       
\usepackage{microtype}      
\usepackage{xcolor}         
\usepackage{amsmath}
\usepackage{graphicx}
\usepackage{xspace}

\newcommand{\jetclass}{{\textsc{JetClass}}\xspace}
\newcommand{\jetclassii}{{\textsc{JetClass-II}}\xspace}

\title{An Evaluation of Representation Learning Methods in Particle Physics Foundation Models}

\author{
  Michael Chen, Raghav Kansal\thanks{Also affiliated with the Fermi National Accelerator Laboratory, Batavia, IL 60510, USA}\thanks{Now at Bexorg, Inc.}, Zichun Hao, Maria Spiropulu
  \\
  Division of Physics, Mathematics and Astronomy \\
  California Institute of Technology \\
  Pasadena, CA 91125 \\
  \\
  \And
  Abhijith Gandrakota, Jennifer Ngadiuba \\
  Particle Physics Division \\
  Fermi National Accelerator Laboratory \\
  Batavia, IL 60510 \\
}

\begin{document}
\begin{flushright}
    FERMILAB-PUB-25-0821-CMS-LDRD-PPD
\end{flushright}

\maketitle

\begin{abstract}
  We present a systematic evaluation of representation learning objectives for particle physics within a unified framework. 
  Our study employs a shared transformer-based particle-cloud encoder with standardized preprocessing, matched sampling, and a consistent evaluation protocol on a jet classification dataset. 
  We compare contrastive (supervised and self-supervised), masked particle modeling, and generative reconstruction objectives under a common training regimen. 
  In addition, we introduce targeted supervised architectural modifications that achieve state-of-the-art performance on benchmark evaluations. 
  This controlled comparison isolates the contributions of the learning objective, highlights their respective strengths and limitations, and provides reproducible baselines. 
  We position this work as a reference point for the future development of foundation models in particle physics, enabling more transparent and robust progress across the community.
\end{abstract}

\section{Introduction}

Precision measurements and searches in high energy physics (HEP) increasingly rely on machine learning (ML) models that operate directly on variable-length sets of particles. 
In this regime, transformer- and point-cloud–based architectures have delivered state-of-the-art (SOTA) performance in classifying jets --- collimated sprays of high energy particles abundant in collisions at the CERN Large Hadron Collider, building on permutation-equivariant encoders and attention with inductive biases tailored to particle interactions \cite{qu_particlenet_2020,qu_particle_2024,zaheer_deep_2018,komiske_energy_2019}. 
Despite rapid progress, empirical comparisons across objectives remain difficult; training protocols, preprocessing choices, sampling strategies, and evaluation metrics often vary from paper to paper, obscuring the contribution of the learning objective itself. 
Public large-scale datasets (e.g., \jetclass) and recent expansions (e.g., \jetclassii) have helped standardize inputs and labels, but consistent end-to-end pipelines and objective-level ablations are still unassessed \cite{li_accelerating_2024}.

Permutation-equivariant encoders and message passing on particle sets underlie many modern jet taggers. 
Early work includes energy flow networks~\cite{komiske_energy_2019}, which leverage infrared (IR) / collinear-safe observables via a deep sets architecture~\cite{zaheer_deep_2018}, and ParticleNet, which operates on jets represented as point or ``particle clouds'' with dynamic-graph convolutions~\cite{qu_particlenet_2020}.
More recently, transformer-based models with class tokens and pairwise-interaction biases, such as the particle transformer (ParT), have set strong baselines across jet-tagging tasks \cite{qu_particle_2024}. 
These works collectively motivate our choice to adopt a ParT-style encoder and to keep the architecture fixed while varying only the learning objective.

In representation learning, contrastive approaches based on InfoNCE/NT-Xent (e.g., SimCLR) and the supervised variant (SupCon) are standard tools \cite{chen_simple_2020,khosla_supervised_2021}. 
Methods that stabilize invariance without negatives, such as VICReg and SimSiam, further expand the design space \cite{bardes_vicreg_2022,chen_exploring_2020}. 
In contrast, masked modeling reconstructs missing content and has proven effective in vision and point-cloud domains \cite{he_masked_2021,yu_point-bert_2022,pang_masked_2022}, while generative objectives such as VAEs learn latent structure via reconstruction with Kullback-Leibler (KL) regularization and remain prevalent \cite{kingma_auto-encoding_2022}. 
In HEP, emerging datasets (\jetclass and \jetclassii) are enabling more standardized comparisons in jet physics \cite{li_accelerating_2024}, but differences in data preprocessing, augmentation techniques, sampling under class imbalance, and metric selection continue to complicate cross-paper synthesis.

We address these gaps by isolating representation learning objectives within a unified framework for particle-cloud jets. 
We fix the encoder family to a transformer backbone with pairwise-interaction attention, apply standardized, physics-informed preprocessing, and evaluate with matched sampling and identical optimization hyperparameters. 
Within this controlled setting, we compare contrastive objectives (self-supervised and supervised), masked particle modeling, and generative reconstruction, and report results on a common benchmark. 
Beyond the objective study, we introduce targeted architectural upgrades that improve fully supervised baselines. 
We provide strong supervised references and help contextualize where self-supervised pretraining is most beneficial for HEP.

\section{Methods}

All methods share a ParT-style particle-cloud encoder to isolate the effect of the learning objective \cite{qu_particle_2024}. 
The encoder embeds per-particle features with a 3-layer MLP (512 hidden, 128 output neurons) and then employs 8 particle-attention blocks and 2 class-attention blocks. Particle-attention blocks are similar to self-attention but augmented with a physics-informed pairwise particle-interaction bias. The pairwise interaction features include $\ln \Delta R$, $\ln k_T$, $\ln z$, and $\ln m^2$, embedded via a 4-layer pointwise 1D convolution with channel dimensions matching the 8 attention heads.
The latent dimensionality is fixed to 128; dropout, LayerNorm and GELU are used in blocks, and the final task head is a linear classifier attached to the pooled latent.
We adopt a scale-invariant per-jet normalization. 
Each jet’s 4-vectors are rescaled such that the jet transverse momentum $p_\mathrm{T}$ equals to $500~\mathrm{GeV}$, following \cite{li_accelerating_2024}. 
Particle features are sanitized to avoid non-physical states. Particles violating physics constraints after augmentation are reverted to their original values, and NaN/Inf values are sanitized throughout.
For objectives that require a projection head, we use a lightweight MLP with ReLU.
Masked modeling and VAE objectives attach a decoder only during pretraining; the supervised head is used only during fine-tuning or for the fully supervised reference model.

\subsection{Learning objectives}

\paragraph{Self-supervised contrastive (JetCLR)} 
We generate two physics-preserving augmented views per jet and train with a normalized temperature-scaled cross-entropy (NT-Xent) loss at temperature $\tau=0.1$ on the projected latent representations. 
Augmentations are based on physical symmetries or experimental invariants such as rotations, translations, and soft and collinear augmentations enforcing infrared and collinear safety, following \cite{dillon_symmetries_2022}.

\paragraph{Masked particle modeling (MPM)}
We randomly mask $\sim$30\% of particles per jet and reconstruct the masked tokens from the encoder output with a small decoder, following \cite{golling_masked_2024}. 
The loss is a sum of per-feature reconstruction terms over masked tokens only.

\paragraph{Generative reconstruction (CLIP-VAE)}
A VAE based on \cite{liu_fast_2023} is trained on the encoder latents with a reconstruction term over particle features and a KL regularizer to a standard normal prior.

\paragraph{Supervised contrastive (SupCon)}
We introduce and adapt the SupCon objective \cite{khosla_supervised_2021} for this task with temperature $\tau=0.07$. 
To increase robustness and enforce invariances, we create two physics‑informed augmented views per jet and form positives from same‑class samples across views. 
Augmentations include mild noise, smearing, dropout, and simple kinematic transforms; validation uses only noise/smearing .

\subsection{Optimization and schedules}

An epoch is defined as $2\,\mathrm{M}$ sampled jets (streamed, not a full sweep). 
Unless noted, we use the AdamW optimizer with learning rate $1\times10^{-4}$, batch size 256, and gradient clipping at 1.0 where applied. Validation is performed every epoch on a balanced $2\,\mathrm{M}$-jet sample with the same sampling policy as training. We pretrain for 50 epochs and fine-tune for 20 epochs. During pretraining we select the encoder by the lowest validation loss and save checkpoints as encoder weights are extracted for fine-tuning; for fine-tuning and supervised baselines we select by highest macro ROC--AUC on the balanced validation set.

\subsection{ParT modifications}
\label{sec:part_mods}
We find the following optimizations of the ParT architecture improve the baseline fully-supervised performance. We replace the feed-forward activation with GEGLU, increase the feed-forward expansion, and add DropPath with small residual scaling to ease optimization. 
We also widen the readout from a single linear layer to a two-layer classifier MLP with GELU and dropout to reduce the pooling bottleneck similar to \cite{li_accelerating_2024}. 
Furthermore, we test the Muon optimizer \cite{noauthor_muon_nodate} under the same schedule and find it to be more performant than AdamW.

\section{Dataset and Experimental Setup}

We use a subset of the \jetclassii dataset with all four-prong categories removed, comprising $bb$, $qq$, $qqb$, $bcs$, three di-tau channels ($\tau_h\tau_e$, $\tau_h\tau_\mu$, $\tau_h\tau_h$), and all $\mathrm{QCD}$ jets. Splits (train/val/test) are disjoint at the file level; the natural test set preserves original class frequencies. 
Training and validation use a stratified sampler that draws $2\,\mathrm{M}$ jets per epoch with
\[
\text{$1/3$ signal}~(bb+qq+\tau\tau),\quad \text{$1/3$}~(qqb+bcs),\quad \text{$1/3$ QCD},
\]
and within the signal bucket, $bb$ at $1/9$, $qq$ at $1/9$, each di-tau channel at $1/27$. 
Jets are represented as variable-length sets of reconstructed particles with per-particle 4-vectors and auxiliary features such as particle type. 
For each objective we (i) pretrain on the training split with the balanced sampler and the objective-specific head, (ii) initialize a 7-way classifier from the pretrained encoder, and (iii) fine-tune on the same training split under the balanced sampler; validation uses the validation split with the same policy. 
The primary metric is macro ROC--AUC; we also report micro ROC--AUC, macro-F1, overall accuracy, and per-class AUCs. 
The primary test set is the natural-distribution test split. 
All experiments use multiple NVIDIA A100 GPUs with the \texttt{PyTorch DistributedDataParallel} module.

\section{Results}
We report results on the natural test distribution. 
The natural test split preserves the original \jetclassii class frequencies and file partitioning where no class reweighting or balancing is applied and evaluation streams over all test files. 
We report per-class one-vs-rest ROC--AUC to measure how well each class separates from the union of all others, which directly reflects multi-class separability and exposes confusions among signal classes.
In addition to the above metrics, we also quantify background rejection at fixed signal efficiencies. 
ParT embeddings refer to the class-token features from our ParT baseline trained with cross-entropy on the classification task.

\begin{table}[h!]
  \caption{Global metrics on the test set.}
  \label{tab:global_metrics}
  \centering
  \resizebox{0.70\linewidth}{!}{
  \begin{tabular}{lcccc}
    \toprule
    Method & Accuracy & Macro AUC & Micro AUC & Macro-F1 \\
    \midrule
    Ours (Supervised) & \textbf{0.865} & \textbf{0.977} & \textbf{0.988} & \textbf{0.834} \\
    Ours (SupCon) & 0.683 & \textbf{0.977} & 0.944 & 0.743 \\
    ParT & 0.820 & 0.975 & 0.981 & 0.799 \\
    CLIP-VAE & 0.094 & 0.818 & 0.571 & 0.176 \\
    JetCLR & 0.082 & 0.797 & 0.578 & 0.156 \\
    MPM & 0.078 & 0.809 & 0.518 & 0.134 \\
    \bottomrule
  \end{tabular}%
  }
\end{table}

\begin{table}[h!]
  \caption{Per-class one-vs-rest ROC--AUC on the test set. Each entry is computed by treating one class as signal and all others as background; the macro AUC is the unweighted mean over classes.}
  \label{tab:perclass_auc}
  \centering
  \resizebox{\textwidth}{!}{
  \begin{tabular}{lccccccc}
    \toprule
    Method & AUC$_{bb}$ & AUC$_{\tau_h\tau_e}$ & AUC$_{\tau_h\tau_\mu}$ & AUC$_{\tau_h\tau_h}$ & AUC$_{qqb/bcs}$ & AUC$_{qq}$ & AUC$_{\mathrm{QCD}}$ \\
    \midrule
    Ours (Supervised) & 0.988 & 0.997 & 0.998 & 0.998 & 0.963 & \textbf{0.937} & \textbf{0.958} \\
    Ours (SupCon) & \textbf{0.993} & \textbf{0.999} & \textbf{0.999} & \textbf{0.999} & \textbf{0.975} & 0.917 & 0.955 \\
    ParT & 0.988 & \textbf{0.999} & \textbf{0.999} & \textbf{0.999} & 0.960 & 0.927 & 0.955 \\
    CLIP-VAE & 0.746 & 0.975 & 0.975 & 0.922 & 0.795 & 0.582 & 0.728 \\
    JetCLR & 0.598 & 0.969 & 0.970 & 0.954 & 0.823 & 0.493 & 0.772 \\
    MPM & 0.757 & 0.962 & 0.963 & 0.919 & 0.772 & 0.588 & 0.700 \\
    \bottomrule
  \end{tabular}%
  }
\end{table}

\begin{table}[h!]
  \caption{Per-class one-vs-rest background rejection ($\mathrm{Rej}=1/\epsilon_B$) at $\epsilon_S=50\%$ on the test set.}
  \label{tab:rej50}
  \centering
  \resizebox{0.80\linewidth}{!}{
  \begin{tabular}{lcccccc}
    \toprule
    Method & Rej$_{bb}$ & Rej$_{\tau_h\tau_e}$ & Rej$_{\tau_h\tau_\mu}$ & Rej$_{\tau_h\tau_h}$ & Rej$_{qqb/bcs}$ & Rej$_{qq}$ \\
    \midrule
    Ours (Supervised)     & \textbf{1030} & 227105   & 145236   & 174360   & 171 & \textbf{65} \\
    Ours (SupCon)   & 758           & \textbf{4428557} & \textbf{4429694} & 512823   & \textbf{218} & 49 \\
    ParT                 & 716           & 1476186  & 885939   & \textbf{622714} & 119 & 56 \\
    CLIP-VAE             & 5             & 86       & 85       & 17       & 6   & 2 \\
    JetCLR               & 3             & 69       & 70       & 38       & 8   & 2 \\
    MPM                  & 5             & 70       & 71       & 18       & 5   & 3 \\
    \bottomrule
  \end{tabular}%
  }
\end{table}

\begin{table}[h!]
  \caption{Per-class one-vs-rest signal efficiency ($\epsilon_S$) at background efficiency $\epsilon_B=10^{-2}$ on the test set.}
  \label{tab:es_1e2}
  \centering
  \resizebox{0.75\linewidth}{!}{
  \begin{tabular}{lcccccc}
    \toprule
    Method & $\epsilon_{S,bb}$ & $\epsilon_{S,\tau_h\tau_e}$ & $\epsilon_{S,\tau_h\tau_\mu}$ & $\epsilon_{S,\tau_h\tau_h}$ & $\epsilon_{S,qqb/bcs}$ & $\epsilon_{S,qq}$ \\
    \midrule
    Ours (Supervised)     & 0.836 & 0.992 & 0.994 & 0.993 & 0.592 & \textbf{0.428} \\
    Ours (SupCon)   & \textbf{0.837} & \textbf{0.999} & \textbf{0.999} & \textbf{0.998} & \textbf{0.626} & 0.390 \\
    ParT                 & 0.790 & 0.997 & 0.997 & 0.997 & 0.529 & 0.406 \\
    CLIP-VAE             & 0.082 & 0.452 & 0.445 & 0.130 & 0.036 & 0.006 \\
    JetCLR               & 0.016 & 0.384 & 0.385 & 0.244 & 0.066 & 0.008 \\
    MPM                  & 0.071 & 0.388 & 0.391 & 0.148 & 0.033 & 0.013 \\
    \bottomrule
  \end{tabular}
  }
\end{table}

\begin{figure}[h!]
  \centering
  \begin{minipage}{0.4\linewidth}
    \centering
    \includegraphics[width=\linewidth]{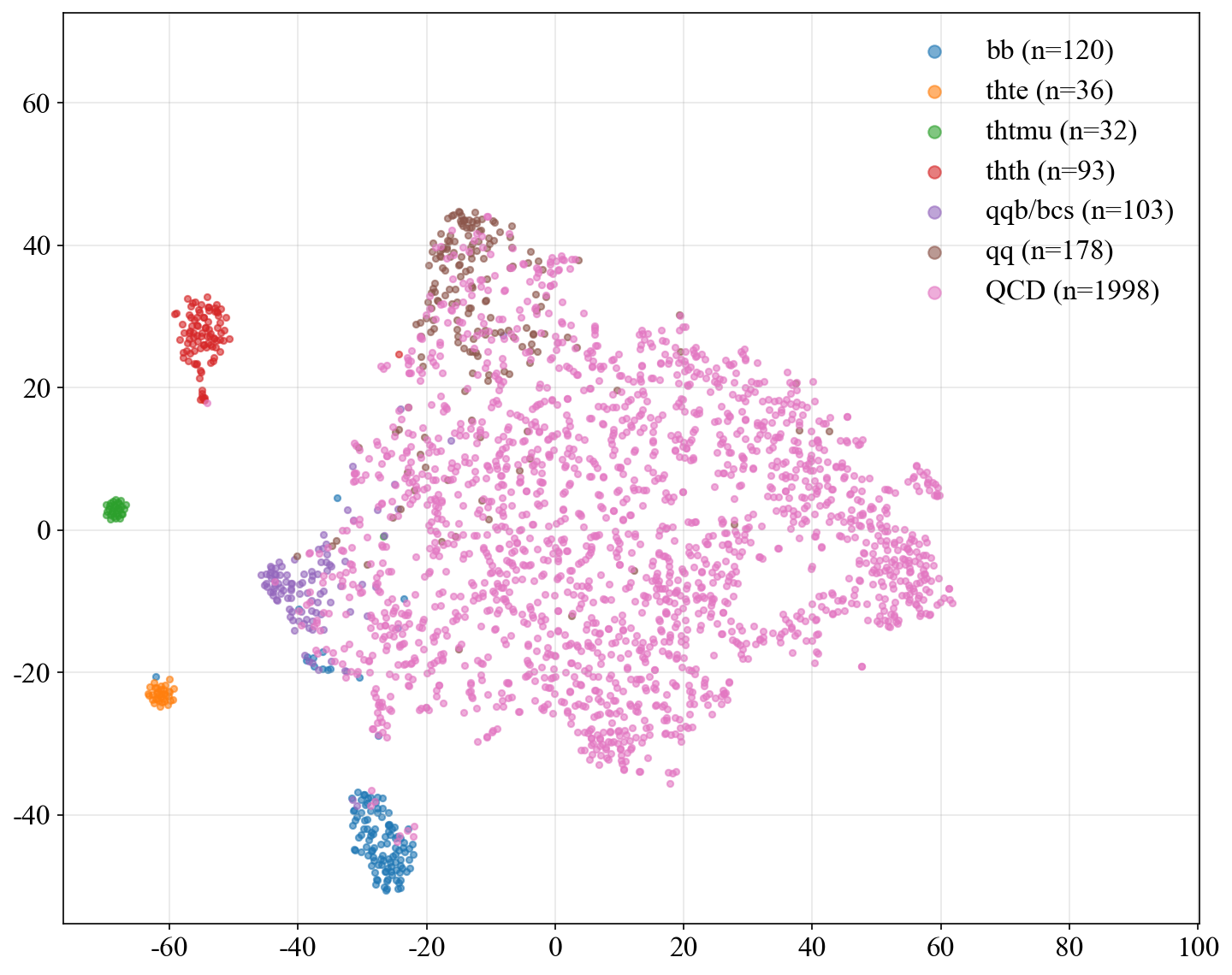}
  \end{minipage}
  \begin{minipage}{0.4\linewidth}
    \centering
    \includegraphics[width=\linewidth]{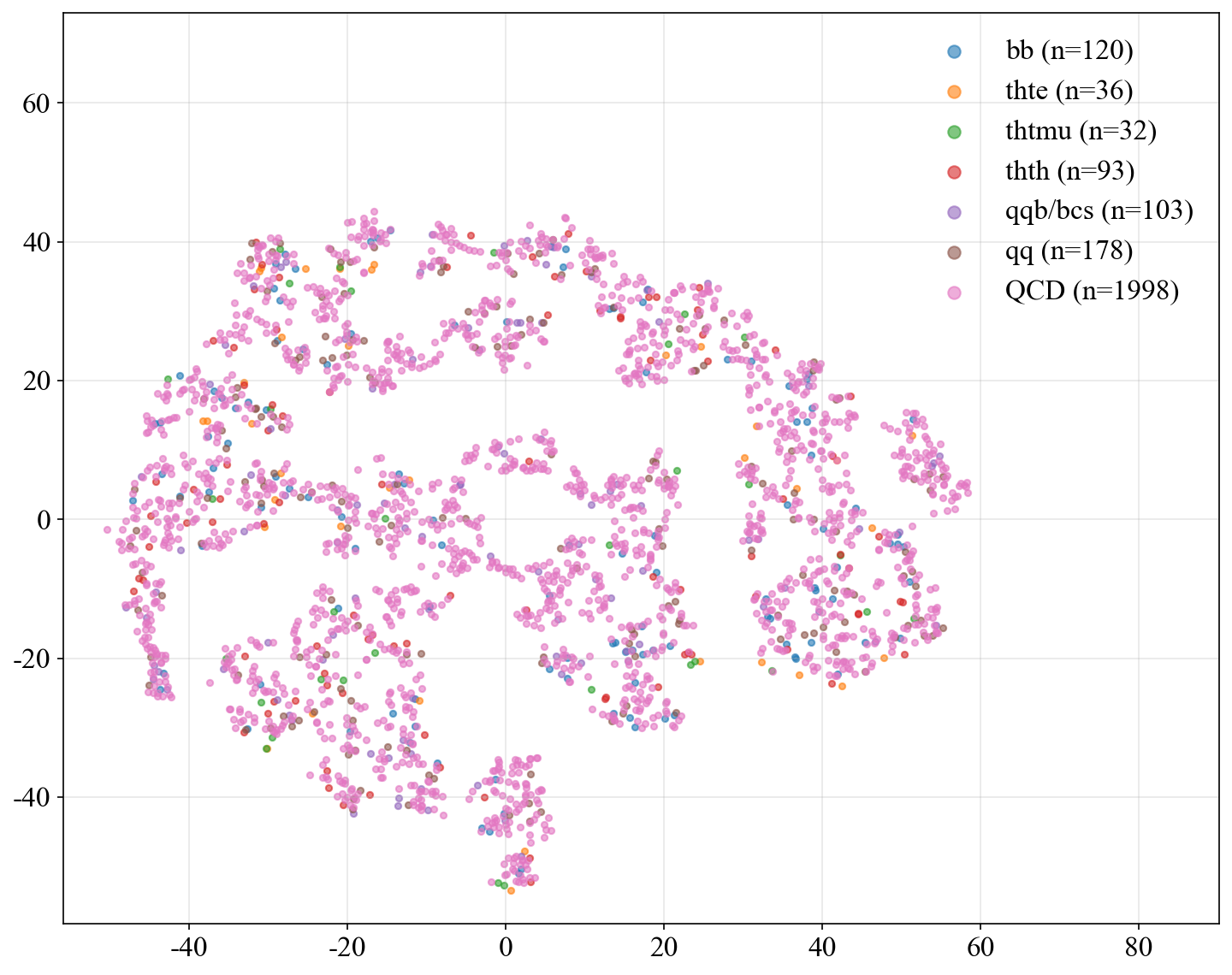}
  \end{minipage}
  \caption{Left: t-SNE of SupCon validation embeddings. Right: t-SNE of our fully-supervised model's embeddings.}
  \label{fig:tsne}
\end{figure}

We find the fully supervised ParT baseline sets a strong reference (Table~\ref{tab:global_metrics}), while the targeted architectural changes described in Sec.~\ref{sec:part_mods} (Ours) further improve accuracy. 
Among representation-learning objectives, SupCon closes most of the gap to supervised training in macro AUC, while SSL methods (JetCLR/MPM/CLIP-VAE) underperform, especially on $qq$ and $\mathrm{QCD}$ jets (Table~\ref{tab:perclass_auc}). 
Fixed-efficiency operating points show large gains in background rejection for our fully-supervised and SupCon models (Table~\ref{tab:rej50}).
Figure~\ref{fig:tsne} illustrates the efficacy of SupCon in producing well-clustered and meaningful embeddings.

\section{Conclusion}
We establish a unified benchmark for objective-level comparisons in particle-cloud representation learning that fixes the transformer encoder, standardizes preprocessing, matches hierarchical sampling, and holds metrics constant. 
Within this benchmark, we carry out a controlled evaluation of contrastive, masked-prediction, and generative objectives for jets. 
Fully supervised training with a ParT-style encoder achieves the best overall accuracy, while supervised contrastive pretraining is the strongest representation-learning objective. 
It matches supervised macro ROC--AUC, mostly leads per-class AUCs, and yields well-structured, semantically meaningful embeddings. Furthermore, we present targeted supervised architectural improvements that achieve strong performance in our setting and serve as reproducible baselines. 
Overall, these results establish a transparent reference point for future foundation models in particle physics, clarifying the role of pretraining objectives disentangled from confounding changes in architectures or training pipelines.

\begin{ack}
This research used resources of the Argonne Leadership Computing Facility, a U.S. Department of Energy (DOE) Office of Science user facility at Argonne National Laboratory and is based on research supported by the U.S. DOE Office of Science-Advanced Scientific Computing Research Program, under Contract No. DE-AC02-06CH11357.
R.K., Z.H., and M.S. are supported by the California Institute of Technology High Energy Physics under Contract DE-SC0011925 with the United States Department of Energy, Office of High Energy Physics. 
A.G. and J.N. are supported by the DOE Office of Science, Award No. DE-SC0023524, FermiForward Discovery Group, LLC under Contract No. 89243024CSC000002 with the U.S. Department of Energy, Office of Science, Office of High Energy Physics, LDRD L2024-066-1, Fermilab, DOE Office of Science, Office of High Energy Physics ``Designing efficient edge AI with physics phenomena'' Project (DE-FOA-0002705), DOE Office of Science, Office of Advanced Scientific Computing Research under the ``Real-time Data Reduction Codesign at the Extreme Edge for Science'' Project (DE-FOA-0002501).
R.K. and J.N are also supported by the AI2050 program at Schmidt Futures (Grant G-23-64934).
\end{ack}

\printbibliography


\end{document}